\documentclass{article}

\usepackage{microtype}
\usepackage{graphicx}
\usepackage{subfigure}
\usepackage{booktabs} 
\usepackage{multirow}
\usepackage{graphicx}

\usepackage[accepted]{icml2024}

\usepackage{amsmath}
\usepackage{amssymb}
\usepackage{mathtools}
\usepackage{amsthm}

\usepackage[capitalize,noabbrev]{cleveref}

\theoremstyle{plain}

\theoremstyle{definition}

\theoremstyle{remark}

\usepackage[textsize=tiny]{todonotes}

\icmltitlerunning{Fine-Tuning Medical Language Models for Enhanced Long-Contextual Understanding and Domain Expertise}

\begin{document}

\twocolumn[
\icmltitle{Fine-Tuning Medical Language Models for Enhanced Long-Contextual Understanding and Domain Expertise}




\begin{icmlauthorlist}
\icmlauthor{Qimin Yang}{yyy}
\icmlauthor{Rongsheng Wang}{yyy}
\icmlauthor{Jiexin Chen}{yyy}
\icmlauthor{Runqi Su}{yyy}
\icmlauthor{Tao Tan}{yyy}

\end{icmlauthorlist}

\icmlaffiliation{yyy}{Faculty of Applied Sciences, Macao Polytechnic University
, Macao, China}
\icmlcorrespondingauthor{Tao Tan}{taotanjs@gmail.com}

\icmlkeywords{Machine Learning, ICML}
\vskip 0.3in]


\printAffiliationsAndNotice{}  

\begin{abstract}

Large Language Models (LLMs) have been widely applied in various professional fields. By fine-tuning the models using domain specific question and answer datasets, the professional domain knowledge and Q\&A abilities of these models have significantly improved, for example, medical professional LLMs that use fine-tuning of doctor-patient Q\&A data exhibit extraordinary disease diagnostic abilities. However, we observed that despite improvements in specific domain knowledge, the performance of medical LLM in long-context understanding has significantly declined, especially compared to general language models with similar parameters. The purpose of this study is to investigate the phenomenon of reduced performance in understanding long-context in medical LLM. We designed a series of experiments to conduct open-book professional knowledge exams on all models to evaluate their ability to read long-context. By adjusting the proportion and quantity of general data and medical data in the process of fine-tuning, we can determine the best data composition to optimize the professional model and achieve a balance between long-context performance and specific domain knowledge.

\end{abstract}

\section{Introduction}

LLMs have demonstrated excellent performance in multiple professional fields, they have exhibited powerful understanding and generation capabilities in natural language processing (NLP) tasks by absorbing massive amounts of general data and specialized domain data. However, despite significant progress in domain-specific knowledge mastery, these models tend to suffer from a decline in long-context understanding and instruction following abilities. This phenomenon has a negative impact on the overall performance of the model in practical applications, limiting its performance in scenarios that require comprehensive understanding and multi-tasking. For example, an LLM fine-tuned with medical data may be able to provide precise medical advice, but may not perform as expected when multiple rounds of conversations are involved or when contextual information is required to understand and answer questions. The normal process of medical diagnosis inquiry should include multiple rounds, coherent, and logical dialogue. Existing medical diagnosis models generally lose track of the previously described symptoms after 2-3 rounds of dialogue, or ignore some key symptoms that have been described in lengthy descriptions, thereby affecting the diagnostic results.

Therefore, it has become an important topic to study how to improve or at least preserve the long-context understanding ability of the model while maintaining professional knowledge capabilities. At the heart of this study is finding a balance that allows models to capture deep expertise without losing their ability to understand and process a wide range of texts. Achieving this goal will not only improve the application effect of LLMs in professional fields, but also expand its applicability in a wider range of practical scenarios. We experimentally evaluate the performance of medically-specific LLMs with different data ratios and quantities, analyze the impact of data composition on model performance, and optimize the data combination strategy for fine-tuning to achieve a balance between medical knowledge and broad language comprehension capabilities.

\section{Related Work}

LLMs are one of the major breakthroughs in the field of artificial intelligence in recent years. These models utilize deep learning techniques, especially neural networks based on the Transformer architecture\cite{vaswani2017attention}, and can be trained on massive amounts of data, thus possessing a powerful ability to generate and understand natural language. The emergence of LLMs marks an important milestone in NLP technology, which not only excels in traditional applications such as generating natural language text, automatic translation, and text summarization\cite{radford2019language}, but also demonstrates great potential in emerging areas such as dialogue systems and knowledge answering and sentiment analysis.

LLMs have demonstrated a wide range of applications in various professional domains, significantly improving productivity and innovation. In the medical field, LLMs can help support physicians in clinical decision-making. By analyzing large amounts of medical literature, electronic health records (EHRs), and the latest research data, LLMs can quickly provide diagnostic advice, treatment plans, and drug recommendations, thus assisting doctors in making more accurate decisions. BenTsao\cite{wang2023bentsao} introduced a method that integrates structured medical knowledge bases to "fine-tune" existing large language models. This method enables the model to refer to authoritative medical information when generating responses, thereby improving the quality and accuracy of the answers. IvyGPT\cite{wang2023ivygpt} proposed a new improvement on fine-tuning training data by mixing real question answering data with generated data for fine-tuning, in order to expand the effective training amount. To better utilize the data extracted by ChatGPT and real data, HuatuoGPT\cite{zhang2023huatuogpt} trained a reward model that aligns the language model with the advantages brought by the two types of data, following the approach of RLAIF (reinforcement learning from AI feedback). Their second-generation model, HuatuoGPT-\uppercase\expandafter{\romannumeral2}\cite{chen2023huatuogpt}, proposed a unified domain adaptation protocol that combines continuous pre-training and fine-tuning stages into a single process. Meanwhile, in terms of doctor-patient communication, LLMs can be used to develop intelligent Q\&A systems and chatbots to help patients answer common health questions, book outpatient appointments and manage personal health information\cite{bai2023qwen}.

\section{Methodology}

\begin{figure}[htbp]
\centering
\includegraphics[width=0.5\textwidth]{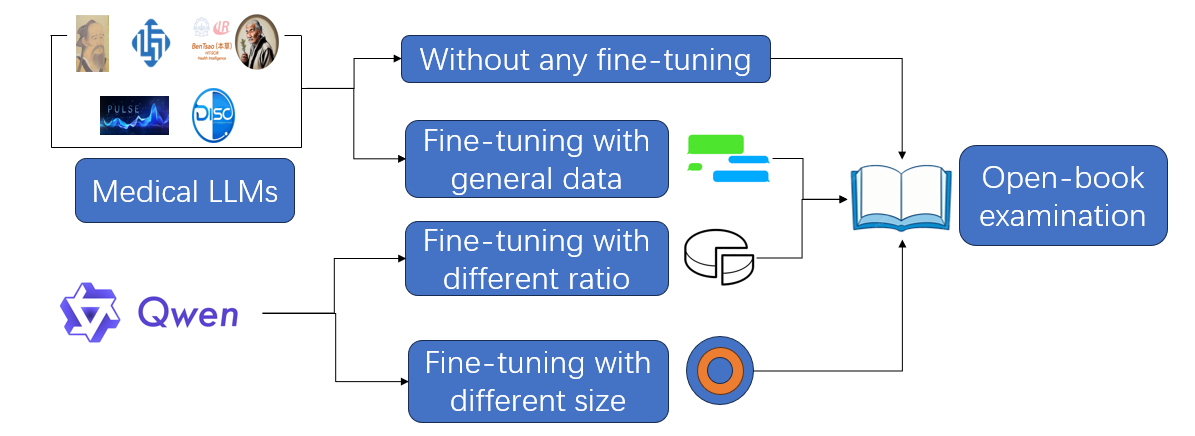}
\caption{The method used in this paper. We conducted relevant experiments on both medical and general models to demonstrate the impact of both general and professional data on the model's ability to understand long contexts. We produce original medical models, medical models that fine-tuned with general data, general model that fine-tuned with different proportions and quantities of data, and finally conducting the same open-book exam to observe their performance.}
\label{fig:method}
\end{figure}

We designed an evaluation method to test the model's contextual capabilities and instruction following capabilities that shown in figure\ref{fig:method}. We have collected some Chinese medical exams, including physician exams, nursing exams, pharmacist exams, medical technology exams, professional knowledge exams and medical postgraduate exams, and use these exams to test the model, all these exams are conducted in a single choice or multiple choice format. But unlike other studies that directly test the model, we input the relevant knowledge required to answer the question as a prompt to the model at the same time as the question, and then require the model to answer only based on the given information and directly output the correct option. 

The relevant fine-tuning datasets we collected include both the publicly available Alpaca Chinese dataset and a large number of Chinese and Western medicine datasets we collected ourselves and sampled from them. Both are Q\&A datasets, with each sample containing instructions, input, and output for supervised fine-tuning. The private dataset consists of collected books, cleaned open source data, Q\&A data from forum websites, and data provided by collaborating hospitals.

\section{Experiments}

\subsection{General Model Exam}
Firstly, we tested some general LLMs that can be publicly used online for daily Q\&A purposes. The results show that the context ability and instruction following ability of the general LLMs are both good, with the majority of testing accuracy reaching over 50\%. Table\ref{table:gen} shows our test results for these general LLMs. Their performance depends not only on the number of parameters and architecture inherent in the model itself, but also on their training methods and data. Models trained with multiple rounds of dialogue or long contextual dialogue can achieve excellent results in this test.

\subsection{Medical Model Exam}
We selected some medical LLMs from the Medical Benchmark in Chinese (CMB)\cite{wang2023cmb} list for the same test, all of which were able to achieve excellent results in the CMB professional medical proficiency test. Table\ref{tab:med} shows our test results for medical LLM. Based on the test results, we found that the results of medical LLM (especially models with better medical capabilities) were relatively unsatisfactory compared to general models. It is worth noting that HuatuoGPT-\uppercase\expandafter{\romannumeral2} has excellent medical abilities among these tested models, but the average accuracy rate of the open book exam is only 4.37\%, far lower than other models, meanwhile, during the process of using PULSE, we found that it often relies on multiple rounds of dialogue to obtain more information for more accurate diagnosis. Long-contextual reading ability and professional ability can be found to be almost a trade-off in the performance of these models. We believe that using a large and extensive proportion of professionals with domain knowledge to fine tune LLM will impair its original ability to understand context and follow instructions. 

\begin{table}
    \centering
    \begin{tabular}{lc}
        \hline
        Model  & Avg.(All exams) \\
        \hline
        GLM-4 & 68.2\% \\
        GPT3.5-Turbo & 41.28\% \\
        GPT4 \cite{achiam2023gpt} & 57.72\%\\
        Qwen-max & 75.76\% \\
        DeepSeek\cite{bi2024deepseek} & 37.52\% \\
        \hline
    \end{tabular}
    \caption{The performance of general LLMs in the six tests mentioned above, Avg. represents the average accuracy they achieved in the tests.}
    \label{table:gen}
\end{table}

\begin{table}
    \centering
    \begin{tabular}{lc}
        \hline
        Model  & Avg.(All exams) \\
        \hline
        PULSE & 34.65\% \\
        DISC-MedLLM\cite{bao2023disc} & 33.36\% \\
        WiNGPT2 & 32.71\% \\
        IvyGPT\cite{wang2023ivygpt} & 32.33\% \\
        HuatuoGPT\cite{zhang2023huatuogpt} & 11.22\% \\
        HuatuoGPT-II\cite{chen2023huatuogpt} & 4.73\% \\
        \hline
    \end{tabular}
    \caption{The average accuracy of medical LLMs in all the medical tests.}
    \label{tab:med}
\end{table}

\subsection{Fine-tuning with general data}
In order to explore the potential improvement of the long-context understanding ability of medical LLMs by general data, we first fine-tune medical LLMs that we tested in the previous section by using the general data that mentioned above. After fine-tuning, we gave these models the same open-book exam as before to evaluate their improvement in long-context understanding and instruction following abilities. Table\ref{tab:med_finetune} shows to what extent our experiment improved model performance. The results indicate that fine-tuning with general data did indeed bring improvements. Means that general data can improve the contextual understanding ability of medical LLMs. This is most evident in HuatuoGPT-\uppercase\expandafter{\romannumeral2}, which basically has the strongest medical ability among these models. It uses a large amount of medical question and answer data for fine-tuning, which makes its ability to understand long-context insufficient. Once it is fine-tuned with general data for a second time, its ability to read long-context can be greatly improved. This also indicates proves that there is almost a trade-off between contextual reading ability and professional ability, and more attention needs to be paid to the use of general data in model fine-tuning. On the contrary, models that have achieved excellent results in previous tests may have a decline issue. When these models were first fine-tuned, they focused on the use of general data and have balanced it well, which made them inherently good at reading long-context content. However, the second fine-tuning actually had a negative effect.

\begin{table}
    \centering
    \begin{tabular}{lcc}
        \hline
        Model  & Avg.(All exams)  & Improvement\\
        \hline
        HuatuoGPT-II & 4.73\%  & +13.5\% \\
        IvyGPT & 32.33\%  & +8.15\% \\
        WiNGPT2 & 32.71\%  & +6.23\% \\
        HuatuoGPT & 11.22\%  & +4.98\% \\
        PULSE & 34.65\%  & -0.15\% \\
        DISC-MedLLM & 33.36\%  & -4.82\% \\
        \hline
    \end{tabular}
    \caption{The results obtained from fine-tuning LLMs in the medical profession using general data and conducting the same open-book testing on them, as well as the improvements obtained before fine-tuning.}
    \label{tab:med_finetune}
\end{table}

\subsection{Fine-tuning with different data composition}
Based on the above knowledge, we believe that general data plays a crucial role in the fine-tuning process of medical professional models. Therefore, we designed a second experiment to investigate the impact of different mixing ratios of general data and medical data on the performance of LLM fine-tuning. The experimental subjects are different parameter versions of the Qwen1.5 model. The reason for using Qwen1.5 is that it has various parameter values to demonstrate the general properties of mixed proportion training in models with different levels of parameters. We have prepared training datasets with different mixing ratios, including a certain proportion of general and medical data. For each parameter size model, we use these mixed datasets for fine-tuning separately. After fine-tuning, we conducted the same medical open-book exam on each model to evaluate their performance in understanding context and following instructions. 

\begin{table}
    \centering
    \begin{tabular}{lcc}
        \hline
        Model  & Avg.(All exams)  & Ratio\\
        \hline
        \multirow{5}*{Qwen1.5-1.8B}& \textbf{39.02\%}  & 9:1 \\
        & 38.67\%  & 4:1 \\
        & 37.58\%  & 1:1 \\
        & 33.38\%  & 1:4 \\
        & 33.73\%  & 1:9 \\
        \hline
        \multirow{5}*{Qwen1.5-4B} & 52.53\%  & 9:1 \\
        & \textbf{52.63\%}  & 4:1 \\
        & 51.14\%  & 1:1 \\
        & 49.96\%  & 1:4 \\
        & 49.51\%  & 1:9 \\
        \hline
        \multirow{5}*{Qwen1.5-7B} & \textbf{62.96\%}  & 9:1 \\
        & 62.27\%  & 4:1 \\
        & 60.42\%  & 1:1 \\
        & 59.96\%  & 1:4 \\
        & 59.65\%  & 1:9 \\
        \hline
        \multirow{5}*{Qwen1.5-14B} & \textbf{68.81\%}  & 9:1 \\
        & 68.53\%  & 4:1 \\
        & 68.1\%  & 1:1 \\
        & 65.12\%  & 1:4 \\
        & 64.85\%  & 1:9 \\
        \hline
    \end{tabular}
    \caption{The average accuracy of models fine tuned using data of different proportions in open-book exams. Ratio represents the ratio of general data to medical professional data.}
    \label{tab:ratio_test}
\end{table}

The result shows in table\ref{tab:ratio_test}, it indicates that a basic trend is that the higher the proportion of general data, the better the performance of each model in the open-book exam. This finding shows the importance of general data in improving the model's long-context understanding and instruction following capabilities, and it is recommended to retain a certain proportion of general data when fine-tuning professional models to maintain comprehensive language understanding capabilities. We believe that the reason for this phenomenon is that data in the medical field is usually more professional, covering relatively narrow content and form, while general question and answer data usually covers a wider range of topics and language forms, including more complex structures and diversity in natural language communication, allowing models to learn a wider range of contextual understanding abilities.

\subsection{Fine-tuning with different data quantity}
Furthermore, within the scope of exploring the impact of data volume effects on specific models in the medical field, we focused on the Qwen1.5-7B model, especially in pure medical data application scenarios, and conducted in-depth analysis. The choice of Qwen1.5-7B as the basic model for research is mainly due to its moderate parameter scale, which facilitates the observation of the direct impact of data volume changes on model performance, and also does not cause over-fitting due to insufficient model parameters or excessive data volume. In order to construct this analytical framework, we systematically prepared a series of medical datasets, gradually expanding the data size from 10k Q\&A pairs to 200k, aiming to discover the trend of the impact of professional data on the model's contextual reading ability. Considering the wide and diverse nature of the medical field, we have specifically included Q\&A pairs from traditional Chinese medicine and Western medicine in our dataset, with a balanced ratio set at 1:1 to address different types of exam questions. Subsequently, using these medical datasets of different scales, we fine tuned the Qwen1.5-7B model and underwent the same open-book examination as before to evaluate the impact of different data volumes on model performance.

\begin{table}
    \centering
    \begin{tabular}{lcc}
        \hline
        Model  & Avg.(All exams)  & Size\\
        \hline
        \multirow{8}*{Qwen1.5-7B}
        & 57.60\%  & 10k \\
        & 55.08\%  & 20k \\
        & 53.64\%  & 50k \\
        & 54.83\%  & 70k \\
        & 51.71\%  & 80k \\
        & 53.41\%  & 90k \\
        & 58.71\%  & 100k \\
        & 56.12\%  & 200k \\
        \hline
    \end{tabular}
    \caption{The average accuracy of the model obtained by fine-tuning with different amounts of data in the open-book exam, where Size represents the number of question and answer pairs used in the fine-tuning dataset. All data are medical professional data, and the ratio of traditional Chinese medicine data to Western medicine data is 1:1.}
    \label{tab:margin_test}
\end{table}

The experimental results shows in table\ref{tab:margin_test} reveal an important phenomenon: in the early stage of limited data size, the increase or decrease in data volume has a significant impact on the context understanding ability and instruction following performance of medical LLM. This means that compared to minor changes in the amount of data used for fine-tuning, the performance of the model will show more significant fluctuations, highlighting the crucial role of data volume at this stage. With the continuous expansion of training data, the overall performance of the model shows a steady upward trend, indicating that the introduction of more samples helps the model learn features more comprehensively, thereby optimizing its performance in complex medical information processing. However, it is worth noting that when the amount of data input reaches a certain threshold, the speed of performance improvement gradually slows down until it reaches a plateau, indicating that the model is beginning to encounter the so-called "data saturation point". This discovery suggests that blindly increasing the amount of data does not always lead to the expected performance leap.

\section{Conclusion}

Through a series of experiments, we explored the impact of the ratio and the amount of general data to medical data on the context understanding and instruction following abilities of medical LLMs. Finding that general LLMs perform well in long-context understanding and instruction following; medical LLMs do not perform satisfactorily in these abilities, but can be improved to some extent through further fine-tuning; the performance of the Qwen1.5 model under different data mixing ratios shows that the general The higher the data ratio, the stronger the model's contextual understanding ability; when using a small amount of medical data, different data volumes have a significant impact on model capabilities, and data volume control is crucial to optimizing the fine-tuning process.

\bibliography{example_paper}
\bibliographystyle{icml2024}

\end{document}